\documentclass[letterpaper,10pt,conference]{ieeeconf}
\IEEEoverridecommandlockouts
\usepackage{amssymb,amsmath,balance,bm,bbm,booktabs,capt-of,caption,graphicx,listings,multicol,multirow,siunitx,stmaryrd,times,trimclip,wrapfig,xcolor,xspace,url,hyperref,placeins}

\definecolor{codegreen}{rgb}{0,0.5,0}
\definecolor{codeblue}{rgb}{0.25,0.5,0.5}
\definecolor{codegray}{rgb}{0.6,0.6,0.6}

\definecolor{pmcolor}{RGB}{70,70,70}

\newcommand{\paragraphc}[1]{\vspace{0.2em}\noindent\textbf{#1}}

\definecolor{haozhipurpole}{HTML}{7A85C1}

\newcommand{\website}{https://choice-policy.github.io}
\definecolor{citecolor}{HTML}{0071bc}
\hypersetup{
    colorlinks=true,
    citecolor=citecolor,
    filecolor=citecolor,
    linkcolor=citecolor,
    urlcolor=citecolor
}
\title{\LARGE \bf Coordinated Humanoid Manipulation with Choice Policies}

\author{
Haozhi Qi$^{*}$, Yen-Jen Wang$^{*}$, 
Toru Lin, Brent Yi, Yi Ma, 
Koushil Sreenath$^{\dagger}$, Jitendra Malik$^{\dagger}$ \vspace{0.5em}\\
UC Berkeley
\thanks{${}^*$ Equal contribution (listed in alphabetical order).}
\thanks{${}^\dagger$ Equal advising.}
}

\begin{document}

\thispagestyle{empty}
\pagestyle{empty}

\let\oldtwocolumn\twocolumn
\renewcommand\twocolumn[1][]{%
\oldtwocolumn[{#1}{
\begin{center}
    \vspace{-1.5em}
    \includegraphics[width=\linewidth]{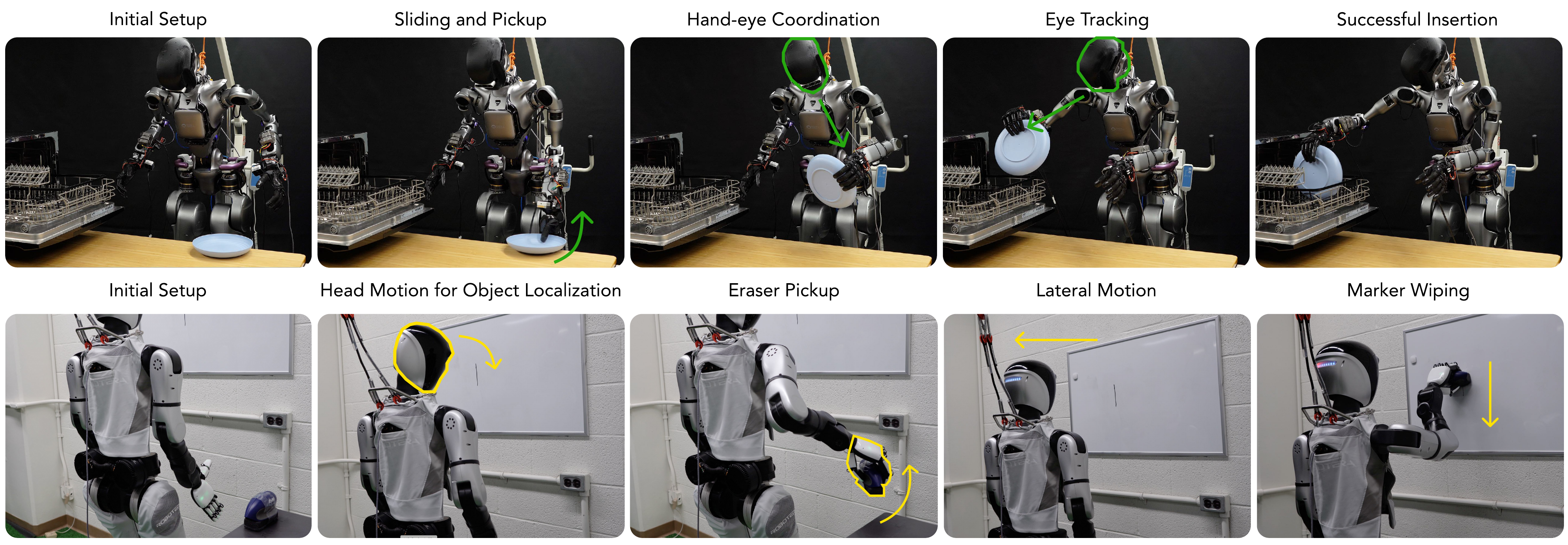}
    \captionof{figure}{\textbf{Coordinated Humanoid Manipulation.} We present a teleoperation system and a policy learning framework for coordinated humanoid manipulation. Our modular interface enables efficient data collection, while the proposed Choice Policy learns to handle multimodal demonstrations with a single forward pass. These components allow the humanoid robot to perform complex, long-horizon tasks, such as dishwasher loading (top row) and whole-body loco-manipulation for whiteboard wiping (bottom row). These tasks require precise coordination between the hands, eyes, and locomotion systems. Videos and additional results are available on \href{\website}{\textcolor{citecolor}{\website}}.}
    \label{fig:teaser}
    \vspace{-0.2em}
\end{center}
}]
}

\maketitle

\begin{abstract}
Humanoid robots hold great promise for operating in human-centric environments, yet achieving robust whole-body coordination across the head, hands, and legs remains a major challenge. We present a system that combines a modular teleoperation interface with a scalable learning framework to address this problem. Our teleoperation design decomposes humanoid control into intuitive submodules, which include hand-eye coordination, grasp primitives, arm end-effector tracking, and locomotion. This modularity allows us to collect high-quality demonstrations efficiently. Building on this, we introduce \textit{Choice Policy}, an imitation learning approach that generates multiple candidate actions and learns to score them. This architecture enables both fast inference and effective modeling of multimodal behaviors. We validate our approach on two real-world tasks: dishwasher loading and whole-body loco-manipulation for whiteboard wiping. Experiments show that Choice Policy significantly outperforms diffusion policies and standard behavior cloning. Furthermore, our results indicate that hand-eye coordination is critical for success in long-horizon tasks. Our work demonstrates a practical path toward scalable data collection and learning for coordinated humanoid manipulation in unstructured environments.
\end{abstract}

\section{Introduction}  

Humanoid robots have the potential to perform complex tasks in human-centric, unstructured environments. To succeed in such settings, they must coordinate their head, hands, and body to actively search for, locate, grasp, and manipulate objects~\cite{sentis2006whole}. Achieving this level of dexterity and flexibility, however, remains highly challenging. It requires seamless whole-body coordination and the tight integration of locomotion and manipulation.

One widely used approach for acquiring new robot skills is learning from demonstrations (LfD). This approach typically involves a human teleoperator controlling the robot’s arms and hands via cameras, motion capture, or Virtual Reality (VR) interfaces to collect expert trajectories, which are subsequently used for supervised imitation learning. However, this approach has several challenges. First, integrating all components is difficult: while some systems incorporate hand-eye coordination~\cite{cheng2024open}, upper-lower body coordination, or waist movement individually, combining them all becomes increasingly complex. Second, existing LfD methodologies often face a trade-off between efficiency and expressiveness: they either rely on iterative computation, which is too slow for the real-time demands of adaptive loco-manipulation, or utilize architectures that are insufficiently expressive to capture the multimodal nature of human demonstrations.

In this paper, we present a system for collecting high-quality demonstrations of coordinated humanoid head, hand, and leg movements. By leveraging this system, we are able to collect high-quality real-world data for complex tasks. We then introduce a novel framework for learning autonomous skills from this dataset. Our method, Choice Policy, captures the multimodal behaviors present in demonstrations without relying on diffusion models or prior tokenization. This approach enables efficient whole-body control through a single forward pass of a neural network.

\paragraphc{Teleoperation Interface.} To overcome the complexity of full-body humanoid control, we developed a modular teleoperation interface designed for both intuitiveness and versatility. Instead of requiring the operator to manage every degree of freedom simultaneously, our system decomposes control into a set of functional submodules. This modularity does not limit the robot's expressive range; instead, it provides a powerful abstraction that allows the user to focus on high-level task logic.

Specifically, our system utilizes a VR controller as the primary input device. The controller’s pose changes are mapped directly to the arm’s end-effector pose. The controller buttons are used to manage atomic grasp types: the four non-thumb fingers are actuated as a single group, while the thumb moves independently. For hand-eye coordination, the user can activate a mode where the head tracks either the left or right hand, reflecting the fact that many manipulation tasks require close hand-object interaction. Finally, the user can use the joystick to command locomotion by leveraging a base policy trained via reinforcement learning (RL) in simulation~\cite{gu2024humanoid,chen2024learning,li2025reinforcement}.

This design provides several advantages. First, the teleoperator can actively select and switch between specific skills instead of continuously mirroring the robot’s full pose, which greatly reduces physical fatigue. Second, by simplifying hand movements into precision and power grasps, the system covers a broad range of manipulation skills while making data collection significantly less demanding. Furthermore, this modular abstraction is not limiting. The interface is designed to be extensible, allowing for the integration of more complex skills such as finger gaiting in the form of additional submodules, as shown in~\cite{lin2024learning,qi2025simple,hsieh2025learning}.

\paragraphc{Policy Learning.} The collected demonstrations are inherently multimodal due to the variations and preferences of the teleoperator. While diffusion policies~\cite{chi2023diffusion} are a popular solution, their inference speed is often too slow for real-time reactive manipulation. These methods frequently require additional optimization~\cite{truong2025beyondmimic,huang2024diffuseloco} or involve a trade-off between fast responsiveness and smooth motion~\cite{black2024pi_0}. Behavior cloning can achieve fast control, but it often struggles to capture the multimodal nature of manipulation. To address this, we propose \textit{Choice Policy}, which is inspired by multi-choice learning literature~\cite{guzman2012multiple}. This approach generates multiple predictions for a given observation, combining the efficiency of a single forward pass with the ability to model multimodality.

During inference, the Choice Policy outputs $K$ candidate action sequences, each with an associated score. The action with the highest score is selected for execution. During training, the score network is supervised to predict the overlap between each proposal and the ground-truth action, which is measured as the negative mean squared error (MSE). The action proposal network is trained using a winner-takes-all paradigm where only the proposal with the smallest MSE is updated through backpropagation. This method enables the policy to capture multimodal behaviors in the dataset while maintaining fast inference speeds.

\paragraphc{Results.} We evaluate our approach on two tasks that highlight the importance of coordinated control for humanoid robots: (1) dishwasher loading and (2) whole-body loco-manipulation for whiteboard wiping. In the dishwasher loading task, the robot’s head must actively shift its gaze from the handover position to the insertion position. In the loco-manipulation task, the robot must maintain stable walking gaits while adapting to errors caused by imprecise initial and final positions.

We conduct comprehensive real-world experiments to study the effects of hand-eye coordination and the proposed Choice Policy algorithm. Our results show that the Choice Policy consistently outperforms both diffusion policy and behavior cloning with action chunking. We further perform ablation studies demonstrating that learned score-based selection is significantly more effective than baselines.

\section{Related Works}

Humanoid locomotion and manipulation have been studied extensively for several decades; we refer the reader to recent surveys for a comprehensive overview~\cite{darvish2023teleoperation,gu2025humanoid}. In this section, we focus our discussion on comparing our approach with existing literature regarding humanoid teleoperation and policy learning.

\subsection{Humanoid Manipulation}

Because humanoids share a similar morphology with humans, a common strategy involves retargeting human movements to robots via keypoint matching. This area has progressed significantly in recent years. ExBody~\cite{cheng2024expressive} decouples upper-body motion tracking from lower-body stability control to demonstrate expressive robot dancing. Subsequent work has further improved motion tracking capabilities by demonstrating agile humanoid movement~\cite{he2025asap,yang2025omniretarget}, extreme balance motion~\cite{zhang2025hub}, and more general reference tracking~\cite{truong2025beyondmimic,ji2024exbody2,chen2025gmt,li2025clone,margolis2025softmimic}.

Combined with advancements in computer vision for human pose estimation, these motion tracking capabilities provide a versatile interface for humanoid teleoperation. Early works such as H2O~\cite{he2024learning}, OmniH2O~\cite{he2024omnih2o}, and HumanPlus~\cite{fu2024humanplus} utilize human keypoints as input and convert them into physically plausible robot control commands learned via sim-to-real transfer. TWIST~\cite{ze2025twist} employs motion capture devices to improve keypoint estimation accuracy, while Sonic~\cite{luo2025sonic} demonstrates system-level improvements that result in smooth and robust behavior.

\begin{figure*}
    \centering
    \includegraphics[width=\textwidth]{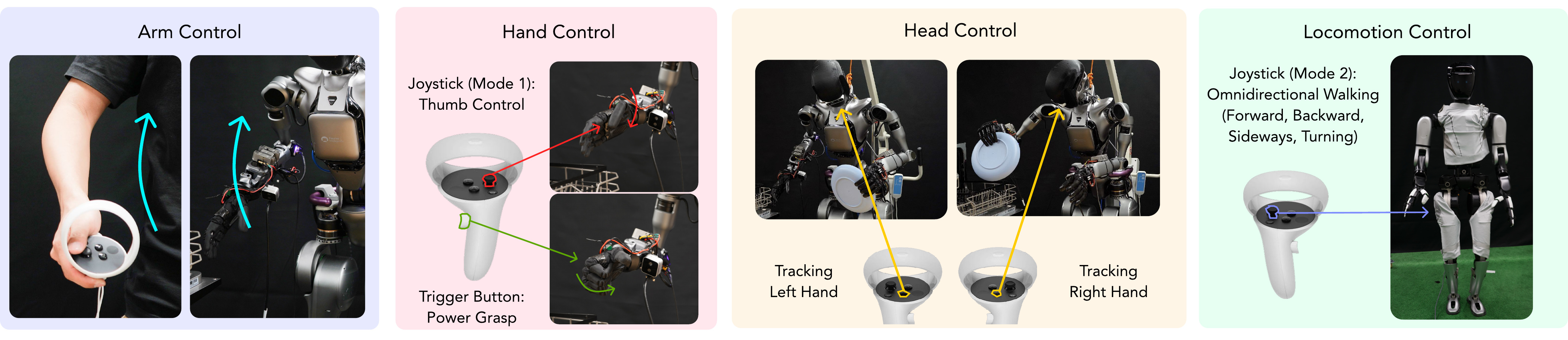}
    \vspace{-0.8em}
    \caption{An overview of our modular teleoperation interface. Control is decomposed into four modules: arm control (end-effector tracking), hand control (trigger button for power/precision grasps and joystick for thumb control), head control (hand-eye tracking of the left or right hand), and locomotion control (omnidirectional walking). A single joystick is shared between thumb control and locomotion; pressing the joystick switches between the two modes. The figure illustrates two humanoid platforms: one mounted on a fixed shelf for manipulation tasks without locomotion, and another with full walking capability. This modular design simplifies data collection while preserving whole-body coordination across arms, hands, head, and legs.}
    \vspace{-1.2em}
    \label{fig:teleop}
\end{figure*}

However, due to retargeting errors, it remains difficult to teleoperate manipulation tasks using only keypoint reference tracking. Alternatively, recent work has explored using Virtual Reality (VR) devices to control humanoid robots by applying whole-body inverse kinematics to the remaining joints, which provides accurate wrist and hand pose information~\cite{ze2024generalizable,lu2025mobile}. Despite this progress, simultaneously controlling all degrees of freedom remains challenging. Most existing systems either control only the upper body or lack active head control. These limitations make data collection difficult, and as a result, learning autonomous policies that can coordinate the head, hands, arms, and legs remains an open problem. For example, Open-Television~\cite{cheng2024open} learn manipulation policies for upper body only and HumanPlus~\cite{fu2024humanplus} learns decoupled upper and lower body motions. While AMO~\cite{li2025amo} demonstrates an autonomous policy for whole-body manipulation, their experiments utilize the Unitree G-1 platform, which is a half-sized humanoid where balancing and control are less demanding. HOMIE~\cite{ben2025homie} also decouples locomotion from upper-body teleoperation, but it utilizes a simpler gripper and demonstrates less complex tasks that lack bimanual coordination or integrated loco-manipulation.

In contrast, our approach provides full-body teleoperation by decomposing coordination into modular sub-skills, including hand-eye coordination, arm tracking, hand grasps, and locomotion. This modularity simplifies both teleoperation and data collection. The resulting high quality demonstrations enable us to train autonomous policies that achieve whole-body manipulation for a full-sized humanoid.

\subsection{Policy Representations}

Imitation learning is a widely used approach for acquiring skills from expert demonstrations. This paradigm includes methods such as simple behavior cloning~\cite{pomerleau1988alvinn} and implicit behavior cloning~\cite{florence2022implicit}. However, due to the nature of human teleoperation, where different operators possess varying preferences, multimodality presents a central challenge. Specifically, for a single observation, multiple expert actions may be valid. Diffusion-based policies~\cite{chi2023diffusion} address this issue by modeling distributions over actions, but their sampling-based inference is often too slow for the real-time requirements of humanoid control~\cite{ze2024generalizable}. Conventional behavior cloning~\cite{torabi2018behavioral,zhao2023learning} offers fast inference through a single forward pass, but it frequently collapses multimodal data into an average, which often leads to suboptimal or unstable actions.

Recent work has attempted to mitigate this limitation by discretizing action spaces or introducing tokenized representations~\cite{shafiullah2022behavior}. However, these methods have not yet demonstrated strong performance on high-degree-of-freedom humanoids that require tight whole-body coordination. In contrast, our approach produces multiple proposals and learns to select among them efficiently. This framework preserves fast inference while capturing the multimodal behaviors necessary for complex manipulation tasks.

\section{Modular Teleoperation Interface}
\label{sec:skilldemo}

Our teleoperation system is illustrated in Figure~\ref{fig:teleop}. Teleoperating a humanoid for precise manipulation is inherently challenging due to the need for whole-body coordination across high-dimensional action spaces. To simplify control, we decompose humanoid manipulation into four modular skills: 1) hand-eye coordination, 2) hand-level atomic grasps, 3) arm end-effector tracking, and 4) omnidirectional walking and standing. While our upper-body design is inspired by HATO~\cite{lin2024learning}, our design of head control and lower-body locomotion introduces novel capabilities for integrated loco-manipulation. Although these modules simplify data collection, they are ultimately unified into a single data-driven policy as described in Section~\ref{sec:choice_policy}.

\paragraphc{Arm Control.} Similar to HATO~\cite{lin2024learning}, arm control is activated only when the trigger button is pressed. We refer to this as \textit{on-demand activation}: at each control step, the VR controller’s relative pose change is transformed into the robot’s frame, from which the absolute end-effector pose is computed. We use inverse kinematics to solve for the target joint positions, which are sent to the humanoid for execution.

This \textit{on-demand activation} is critical for collecting large-scale, precise demonstrations. Many complex tasks require the two arms to operate sequentially rather than simultaneously. Our design allows one arm to remain stationary while the other operates. This prevents the idle arm from drifting or hovering in midair, which reduces teleoperator fatigue and avoids unnecessary robot motion. Additionally, on-demand activation enables the robot to achieve larger pose changes in position or orientation. The human operator can reset and extend the controller iteratively to re-center their workspace. This allows for extended movements without needing to physically reproduce the entire motion. While such selective activation is common in industrial arm teleoperation, it has rarely been adopted for humanoid platforms.

\paragraphc{Hand Control.} The four non-thumb fingers are grouped and actuated together via the grip button, while the joystick independently controls the thumb. Both the grip button and the joystick provide continuous-valued signals that are mapped to finger actuation. This allows the operator to perform fine-grained movements and adjust the tightness of a grasp. Despite this dimensionality reduction, the essential grasp taxonomy is preserved; examples include power grasps, precision grasps, and flattening~\cite{feix2015grasp}. While we currently use these atomic primitives, the framework is not limiting. These skills can be further extended to more complex behaviors. For instance, finger gaiting could be achieved by mapping additional controller inputs to specific finger coordination patterns~\cite{hsieh2025learning}.

This design offers two key advantages for high-quality data collection. First, it simplifies control because maintaining a stable button press is far easier for the teleoperator than maintaining a precise finger pose. Second, it improves accuracy. High-DoF finger tracking often produces jittery and unstable motions, whereas our reduced mapping yields smoother and more reliable grasps.

\paragraphc{Hand-Eye Coordination.} Most manipulation tasks are hand-centric. They require the head to maintain visibility of the active hand. For example, when the robot loads a dishwasher and inserts a plate into the dishrack, it must look at the hand to confirm that the plate is correctly aligned. Motivated by this observation, we implement a button-triggered tracking mode. Pressing the button switches the head to follow either the left or the right hand.

Specifically, let $p_h \in \mathbb{R}^3$ denote the 3D position of the selected hand in the robot base frame, and let $p_{head} \in \mathbb{R}^3$ denote the head position. The relative displacement vector is $r=p_h-p_{head}$. From this vector, we compute the desired yaw and pitch angles that orient the head toward the hand. The yaw is obtained as the azimuth angle in the $xy$-plane: $\text{yaw} = \arctan2(r_y, r_x)$, while the pitch is given by the elevation angle:  $\text{pitch} = \arctan2(-r_z, \sqrt{r_x^2 + r_y^2})$. The roll angle is fixed to zero. Finally, each angle is clipped to its corresponding joint limit before being sent as a target command. This implementation ensures that the head camera continuously points toward the selected hand. Consequently, the manipulation region remains within the field of view, which reduces occlusions during long-horizon tasks.

\paragraphc{Locomotion Policy.} For tasks that require lower-body movement, we use a velocity conditioned reinforcement learning (RL) policy trained in simulation and transferred to the real robot. The policy runs at 100\,Hz and takes velocity commands such as \emph{stand}, \emph{walk forward}, \emph{walk backward}, \emph{walk sideways}, and \emph{turn}, each of which is parameterized by a target velocity. The RL policy outputs target joint positions that are tracked by low-level PD controllers on the robot. During teleoperation, we utilize the Quest controller's joystick to toggle between locomotion and manipulation modes. In locomotion mode, joystick movements specify velocity commands to control the walking direction. In manipulation mode, the joystick independently controls the thumb movement. Releasing the joystick while in locomotion mode defaults the robot to the \emph{stand} action. The training setup, including the reward design, domain randomization, and network architecture, follows~\cite{gu2024advancing}; detailed implementation are provided in our supplementary material.

\paragraphc{Humanoid Platform.} We use the Fourier GR-1 and Robotera Star-1 as examples to demonstrate the generality of our method. The GR-1 is a human-sized robot with a total of 32 actuated motors: 6 motors on each leg, 7 motors on each arm, 3 motors on the waist, and 3 motors on the head. It has two Psyonic Ability Hands mounted; each Ability Hand has 6 motors. In total, this system has 44 degrees of freedom (DoF), providing the flexibility to accomplish various tasks. The Star1 is another humanoid platform, featuring a head with 2 DoF, a waist with 3 DoF, 7 DoF per arm, and 6 DoF per leg. In addition, it is equipped with two XHand, each offering 12 actuated DoF without passive joints. Altogether, the Star1 has 55 actuated DoF, enabling complex whole-body motions and dexterous manipulation.

The above teleoperation primitives are designed primarily to ensure accurate demonstrations. Adopting a modular approach offers several key benefits. First, the reduced complexity makes teleoperation much easier to learn: in our experiments, users typically required less than 10 minutes of practice before they could smoothly perform complex, long-horizon tasks. Second, modularity allows us to balance usability during data collection with flexibility during learning. Although upper-body teleoperation is decomposed into modules, the final policy is a unified neural network that integrates all signals in a fully data-driven manner.  

\section{Choice Policy}
\label{sec:choice_policy}

\begin{figure}
    \centering
    \includegraphics[width=\linewidth]{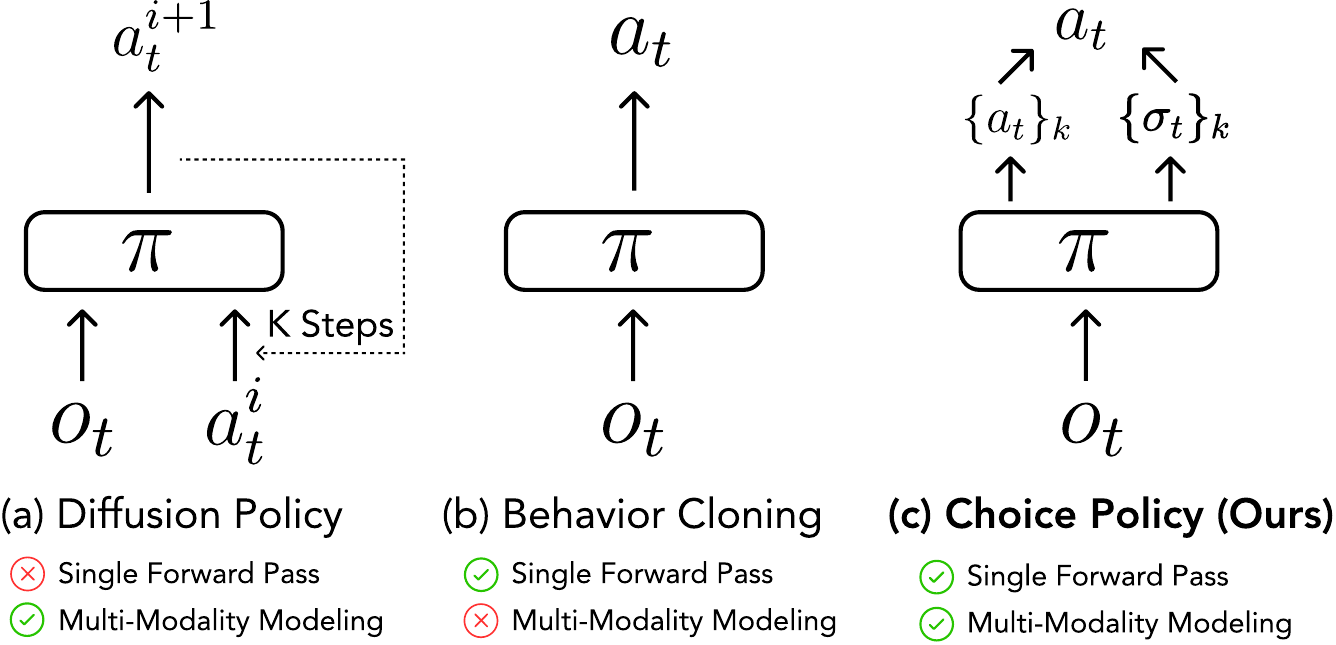}
    \caption{\textbf{Policy architectures for imitation learning.} 
    (a) Diffusion Policy models multi-modality but requires $K$ iterative sampling steps, which makes inference slow. 
    (b) Standard Behavior Cloning enables fast inference with a single forward pass but struggles to capture multimodal behaviors. 
    (c) \textbf{Choice Policy (ours)} combines the strengths of both: it generates $K$ candidate actions in a single forward pass and uses a learned score to select the best one. This design achieves fast inference while effectively handling multi-modality.}
    \vspace{-1em}
    \label{fig:network}
\end{figure}

An overview of our method is shown in Figure~\ref{fig:network}. Given an observation $o_t$, the policy outputs a set of action proposals $\{a_t^{(k)}\}_{k=1}^K$ together with corresponding scores $\{\sigma_t^{(k)}\}_{k=1}^K$. The final action is selected by choosing the proposal with the highest score. During training, the scores are learned to predict the negative mean squared error (MSE) between each candidate trajectory and the ground truth.

The policy takes as input an RGB image and proprioceptive signals; it outputs target joint positions for the arms and hands. When locomotion is involved, the policy additionally outputs high-level commands for the lower body.

\subsection{Background}

Imitation learning via behavior cloning (BC) is a common approach for acquiring robot skills. Given a dataset $\mathcal{D}=\{(o_t, a_t)\}$ of observations and demonstrated actions, the policy $\pi$ is trained to minimize a regression loss, typically the mean squared error (MSE):
$$\mathcal{L}_{\text{BC}} = \| \pi(o_t) - a_t \|^2.$$

While effective, this approach struggles when demonstrations are multimodal. In teleoperation data, humans rarely repeat identical trajectories; therefore, multiple different actions may be valid for the same state. Naively minimizing MSE with a deterministic policy usually causes the model to average across these actions, often producing unrealistic or suboptimal behaviors.

To address this, diffusion policies~\cite{chi2023diffusion} and their variants train generative models that better capture multimodal distributions. This results in a more expressive policy class, but it comes at the cost of slow inference, making real-time control difficult. Although alternative methods exist~\cite{zhang2025flowpolicy}, they typically involve a trade-off between accuracy and computational speed.

In this section, we present an alternative algorithm motivated by multiple-choice learning for structured prediction problems~\cite{guzman2012multiple,guzman2014efficiently}. These problems require a model to output multiple valid hypotheses given a single input. In computer vision, this approach is utilized by SAM~\cite{kirillov2023segment}, where a network predicts multiple possible segmentation masks for a single user click to resolve spatial ambiguity. Similarly, learning from teleoperation data is inherently ambiguous; a single visual observation often corresponds to several valid expert behaviors in the training data. Motivated by these designs, our approach predicts multiple candidate action trajectories along with a score that ranks them, effectively capturing this behavioral diversity without the computational overhead of iterative sampling.

\subsection{Policy Architecture.}

Our model consists of a feature encoder, an action proposal network, and a score prediction network. The feature encoder maps visual observations and proprioceptive inputs into a single vector representation. The action proposal network then takes this feature and predicts a set of candidate action trajectories. In parallel, the score prediction network takes the same feature as input and outputs a score for each trajectory.

\paragraphc{Observation Encoder.} We use two types of observations: visual inputs (RGB and depth) and proprioceptive signals. For RGB, we adopt a frozen DINOv3 feature encoder~\cite{simeoni2025dinov3}. For depth, we use a randomly initialized ResNet-18 trained from scratch, with all batch normalization layers replaced by group normalization as in~\cite{chi2023diffusion}. For proprioception, we use a 3-layer MLP with ReLU activations. Not all tasks use the same set of observation modalities, but all comparisons and baselines are given access to the same information with the same encoder design. The features from different modalities are then concatenated to form the final feature vector $\mathbf{f}_t$. 

\paragraphc{Action Proposal Network.} We then pass the feature vector $\mathbf{f}$ to an action proposal network. Different from typical action prediction, we predict a set of candidate actions $\{\mathbf{a}_k\}_{k=1}^{K}$. Each $\mathbf{a}$ is a multi-step prediction, following the practice of~\cite{chi2023diffusion,zhao2023learning}, therefore $\mathbf{a}_t \in \mathbb{R}^{T \times |A|}$, where $|A|$ is the action dimension. In practice, we use a two-layer MLP with an output channel size $|K| \times |T| \times |A|$ and reshape it to the desired dimension.

\paragraphc{Score Prediction Network.} To select the best action, we must also predict the quality of each candidate trajectory. Similar to~\cite{kirillov2023segment}, which predicts the score of each segmentation by estimating its intersection-over-union (IoU) with the ground truth, our approach predicts the MSE between each action trajectory and the ground-truth trajectory. We implement this using a two-layer MLP that outputs $|K|$ channels, where each channel represents the score of one action trajectory.

\lstset{
  backgroundcolor=\color{white},
  basicstyle=\fontsize{7.5pt}{8.5pt}\fontfamily{lmtt}\selectfont,
  columns=fullflexible,
  breaklines=true,
  captionpos=b,
  commentstyle=\fontsize{8pt}{9pt}\color{codegray},
  keywordstyle=\fontsize{8pt}{9pt}\color{codegreen},
  stringstyle=\fontsize{8pt}{9pt}\color{codeblue},
  frame=tb,
  otherkeywords = {self},
}
\begin{figure}[t]
\tiny
\begin{lstlisting}[language=python]
  # N: batch size
  # K: number of proposals
  # T: action prediction horizon
  # x: input features with shape [N, K, A, T]
  scores = score_pred(x)  # [N, K] 
  actions = action_proposal(x)
  score_idx = scores.argmin(dim=1)
  if train:  # training
    gt = gt[:, None].repeat(1, K, 1, 1)  # [N, K, A, T]
    loss = mse(gt, pred_action).mean(dim=(2, 3))
    score_loss = ((scores - loss.detach()) ** 2).mean()
    action_loss = loss[torch.arange(loss.shape[0]), loss_mask].mean()
    loss = action_loss + score_loss
    return loss
  else:  # inference
    actions = actions[torch.arange(N), score_idx]
    return actions
\end{lstlisting}
\vspace{-0.5em}
\caption{PyTorch pseudocode for the Choice Policy training and inference procedure. We include this snippet to highlight that our method is simple to implement; it requires only a few lines of code to combine winner-takes-all action learning with score regression. This ease of implementation makes the approach readily reproducible.}
\vspace{-1em}
\label{fig:code}
\end{figure}

\subsection{Training Objective}
During training, the policy outputs $K$ candidate action trajectories 
$\{a_t^{(k)}\}_{k=1}^K$ and a set of predicted scores 
$\{\sigma_t^{(k)}\}_{k=1}^K$. 
We first compute the mean squared error (MSE) between each proposal and the ground-truth trajectory $a_t$, averaged over action dimensions and the prediction horizon:
\[
\ell^{(k)} = \frac{1}{|A||T|} \sum_{i=1}^{|A|} \sum_{j=1}^{|T|} 
\big(a_t^{(k)}[i,j] - a_t[i,j]\big)^2.
\]
These per-proposal errors $\{\ell^{(k)}\}$ serve two purposes. First, they are used as regression targets for the score prediction network:
\[
\mathcal{L}_{\text{score}} = \frac{1}{K} \sum_{k=1}^K 
\big(\sigma_t^{(k)} - \ell^{(k)}\big)^2.
\]
Second, we adopt a winner-takes-all strategy for updating the action proposal network. The proposal with the smallest error is selected: $k^\star = \arg\min_{k} \ell^{(k)}$, and only this winning trajectory $a_t^{(k^\star)}$ receives gradients. The corresponding action loss is $\mathcal{L}_{\text{action}} = \ell^{(k^\star)}$.

The final training loss combines these two terms:
$\mathcal{L} = \mathcal{L}_{\text{action}} + \mathcal{L}_{\text{score}}$. This encourages the proposal network to generate diverse candidates while the score network learns to approximate their true quality. This enables the policy to efficiently select the best trajectory during inference time.

\begin{table}[!t]
\centering
\captionof{table}{\textbf{Main results on the dishwasher loading task.} Our Choice Policy achieves superior performance compared to behavior cloning and diffusion policy. Additionally, without hand-eye coordination, none of the methods can reliably complete the full task.}
\setlength{\tabcolsep}{4pt}
\renewcommand{\arraystretch}{1.25}
\resizebox{\linewidth}{!}{%
\begin{tabular}{rcrrr}
\toprule
Method & Hand-eye Coor. & Pickup & Handover & Insertion \\
\cmidrule(r){1-1}
\cmidrule(r){2-2}
\cmidrule(r){3-5}
Diffusion Policy & & 10 / 10 & 8 / 10 & 1 / 10 \\
Behavior Cloning & & 9 / 10 & 6 / 10 & 1 / 10 \\
Choice Policy & & 10 / 10 & 7 / 10 & 2 / 10 \\
\midrule
Diffusion Policy & \checkmark & 10 / 10 & 7 / 10 & 5 / 10 \\
Behavior Cloning & \checkmark & 9 / 10 & 7 / 10 & 5 / 10 \\
Choice Policy & \checkmark & 10 / 10 & 9 / 10 & 7 / 10 \\
\bottomrule
\end{tabular}
}%
\vspace{0.2em}
\label{tab:dishwasher}
\vspace{-1.5em}
\end{table}

\subsection{Inference}

At inference time, the policy generates $K$ candidate trajectories in a single forward pass. The score prediction network assigns a score to each proposal, and the trajectory with the lowest predicted error is selected for execution:
\[
a_t = a_t^{(k^\star)}, \quad 
k^\star = \arg\min_{k} s_t^{(k)}.
\]
This procedure allows the policy to efficiently select among multiple valid behaviors. It preserves the fast inference speed of behavior cloning while effectively handling multi-modality. To demonstrate that our approach is simple to implement, we provide the PyTorch pseudocode in Figure~\ref{fig:code}.

\section{Experiments}

We evaluate our approach on two tasks: (1) \textbf{Dishwasher Loading} and (2) \textbf{Loco-Manipulation Wiping}. The dishwasher loading task is used to test the effectiveness of our learning algorithm, the quality of the teleoperation system, and the importance of hand-eye coordination, as well as to conduct ablation studies. The more challenging loco-manipulation wiping task serves as a demonstration of the flexibility of our teleoperation system and shows that our policy learning methods can extend to more complex, long-horizon tasks.

\subsection{Dishwasher Loading}

\paragraphc{Task Setup.} At the start of each trial, a plate is placed near the edge of a table in front of the robot. Because the plate is thin and cannot be grasped directly from the top, the robot must leverage environmental contacts. It first slides the plate closer to the edge to avoid finger-table collisions and then grasps it securely. The plate is subsequently handed over from one hand to the other before being inserted into the dishrack inside the dishwasher. Since the dishwasher is not initially visible from the head camera and the wrist camera becomes occluded by the plate, the robot must actively adjust its head to track the manipulating hand. Without proper head control, the dishrack cannot be observed.

This task is particularly challenging for two reasons. First, the thin plate requires the use of environmental support, such as sliding along the table surface, to achieve a stable pickup. Second, the task is long-horizon and composed of three tightly coupled sub-skills: pickup, handover, and insertion. In this sequence, failure at any stage leads to overall failure. These factors underscore the need for reliable coordination across subsystems and robust handling of multi-modality in demonstrations.

We demonstrate this task using the Fourier GR-1 humanoid robot. To isolate the effect of hand-eye coordination, we minimize the influence of locomotion and lower-body movement by mounting the robot on a fixed shelf during this experiment. We collect 100 demonstrations for this task.

\begin{figure}
    \centering
    \includegraphics[width=\linewidth]{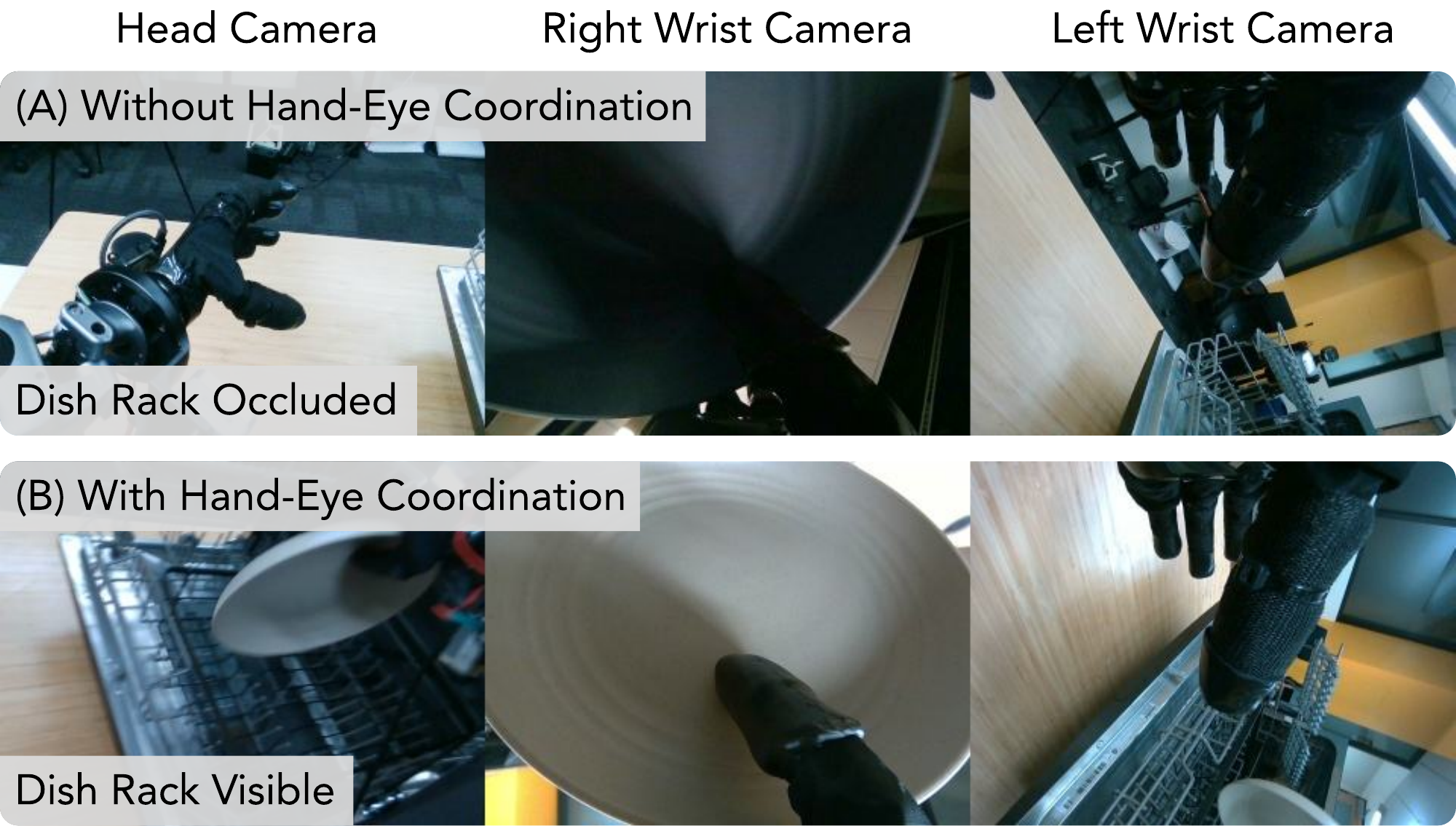}
    \caption{\textbf{Camera view comparison: (A) without hand-eye coordination vs. (B) with hand-eye coordination.} The figure shows head and wrist camera perspectives during the insertion stage. During insertion, the head camera with hand-eye coordination provides a clear view of the dishrack. By contrast, without coordination, the view is occluded, which makes it difficult to verify whether the plate has been successfully inserted.}
    \vspace{-2em}
    \label{fig:handeye}
\end{figure}

\paragraphc{Inputs.} In this task, we use the RGB camera on the robot’s head as well as two cameras mounted on the wrists. We also include the joint positions of the upper body and hands, which results in a 29-dim proprioceptive input vector.

\paragraphc{Main Results.} Table~\ref{tab:dishwasher} compares our method with behavior cloning (BC) and Diffusion Policy (DP). All approaches perform reliably during the pickup stage. However, both BC and DP succeed in only about 70\% of handovers, reflecting the difficulty of modeling multiple valid handover strategies. Their performance drops further during insertion, with success rates of only 50\%. In contrast, Choice Policy maintains high reliability throughout the sequence; it improves handover success to 90\% and achieves 70\% on insertion. These results indicate that Choice Policy handles multi-modality in demonstrations more effectively while retaining fast inference, leading to more consistent execution of the complete task.

\paragraphc{Hand-Eye Coordination.} We also find that head orientation has a strong impact on task performance. Table~\ref{tab:dishwasher} shows results for cases where the head camera does not track the active hand. Without adaptive head motion, all methods still succeed in pickup but struggle in later stages. This is particularly evident during insertion, where success rates drop to nearly zero. This failure occurs because the fixed viewpoint often occludes the dishrack, preventing precise placement. By contrast, with adaptive hand-eye coordination, the robot maintains visibility of the manipulating hand throughout the task, which substantially improves performance in both handover and insertion.

Figure~\ref{fig:handeye} further illustrates this effect: without coordination, the dishrack is occluded, whereas with coordination, it remains clearly visible. These results highlight that hand-eye coordination is critical for complex manipulation tasks, especially when accurate placement depends on maintaining clear visual feedback.

\paragraphc{Out-of-Distribution (OOD) Evaluations.} We further evaluate robustness under out-of-distribution conditions (Table~\ref{tab:dishwasher_ood}) across two settings: \textbf{Color OOD} and \textbf{Position OOD}. In the Color OOD setting, the robot manipulates a green plate, even though only pink, blue, and brown plates were seen during training. In the Position OOD setting, the plate is placed slightly outside the training range.

In both settings, performance drops compared to in-distribution tasks. Color variation mainly reduces insertion accuracy, while position shifts prove to be more challenging. All methods degrade significantly in the latter case; for instance, insertion success falls near zero for diffusion policy and behavior cloning. In contrast, choice policy demonstrates greater robustness by achieving higher success rates in both handover and insertion, though performance still falls short of in-distribution levels. The results highlight the difficulty of generalizing to unseen appearances and positions, while showing that our choice policy maintains comparatively strong performance.

\begin{table}[!t]
\centering
\captionof{table}{Out-of-distribution (OOD) evaluation on the dishwasher loading task. We consider two settings: \textbf{Color OOD}, where a green plate unseen during training is used, and \textbf{Position OOD}, where the plate is placed slightly outside the training range. All methods show performance degradation compared to in-distribution tasks. However, Choice Policy demonstrates greater robustness and achieves higher success rates in both handover and insertion. Results are reported as successes out of 10 trials.}
\setlength{\tabcolsep}{2pt}
\renewcommand{\arraystretch}{1.25}
\resizebox{\linewidth}{!}{%
\begin{tabular}{rrrrrrr}
\toprule
\multirow{2}{*}{Method} & \multicolumn{3}{c}{Color OOD} & \multicolumn{3}{c}{Position OOD} \\
& Pickup & Handover & Insertion & Pickup & Handover & Insertion\\
\cmidrule(r){1-1}
\cmidrule(r){2-4}
\cmidrule(r){5-7}
Diffusion Policy & 9 / 10 & 5 / 10 & 1 / 10 & 6 / 10 & 4 / 10 & 0 / 0\\
Behavior Cloning & 8 / 10 & 7 / 10 & 5 / 10 & 8 / 10 & 2 / 10 & 2 / 10 \\
Choice Policy & 10 / 10 & 9 / 10 & 5 / 10 & 7 / 10 & 4 / 10 & 4 / 10 \\
\bottomrule
\end{tabular}
}
\vspace{-1em}
\label{tab:dishwasher_ood}
\end{table}

\paragraphc{Ablation Experiments.} Since our method outputs $K$ candidate actions at each step, it is natural to ask how performance changes if these proposals are used differently. We evaluate three alternatives: (1) \textbf{Random Choice}, where one proposal is selected uniformly at random; (2) \textbf{Mean Choice}, where the $K$ actions are averaged; and (3) \textbf{Single Choice}, where a single proposal is used consistently across all steps. For the single-choice variant, we report both the best and worst performing heads.

As shown in Table~\ref{tab:dishwasher_ablation}, all variants underperform compared to our scoring-based selection. Random choice achieves moderate pickup success but fails in later stages. Averaging proposals performs poorly because multimodal actions cannot be meaningfully averaged. Furthermore, using a fixed single head is inconsistent and results in large performance gaps between the best and worst cases. In contrast, our scoring-based method achieves high success across all stages. These results confirm that \textbf{the scoring mechanism is critical}, as it allows the policy to adaptively select the most suitable proposal at each step.

\begin{figure*}
    \centering
    \includegraphics[width=\textwidth]{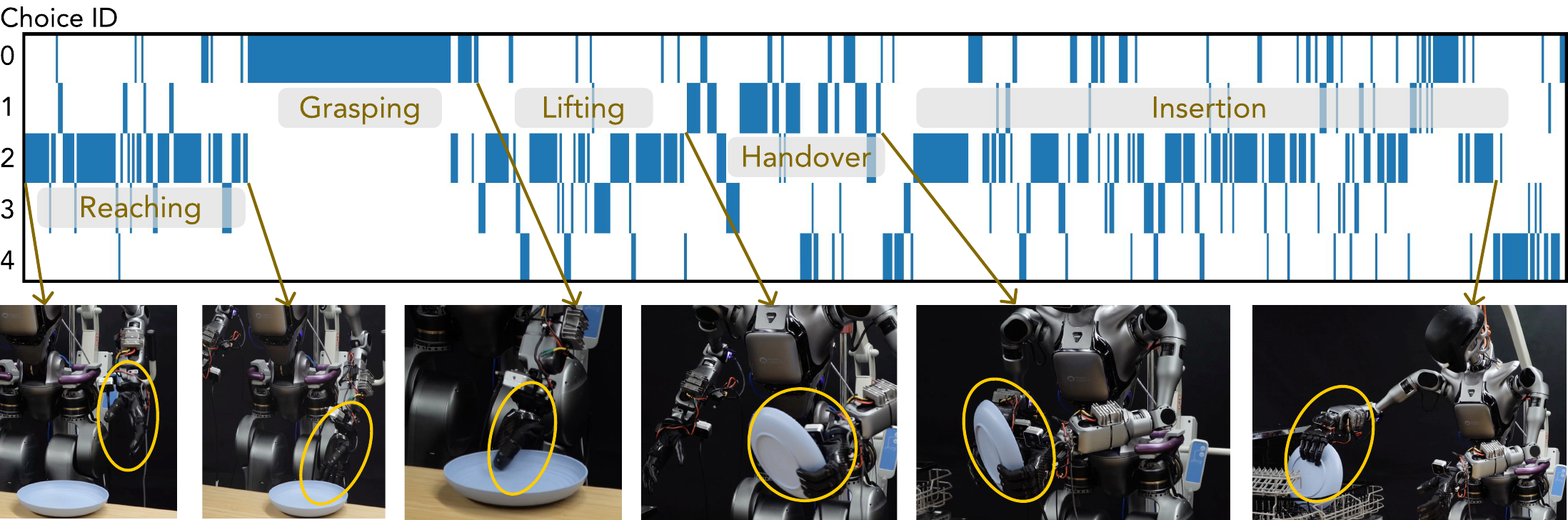}
    \caption{\textbf{Visualization of policy specialization and switching.} We visualize the selection of action proposals during a complete rollout of the dishwasher loading task. Each blue row corresponds to one of the $K=5$ action proposals (Choice IDs 0-4), while the columns represent different task phases. The figure illustrates how different proposal heads specialize in specific sub-skills. At inference time, the score prediction network selects the most suitable expert head for the current phase, which enables the policy to maintain high precision across the entire long-horizon task.}
    \label{fig:choice_vis}
    \vspace{-1em}
\end{figure*}

\paragraphc{Qualitative Visualization.} To better understand the internal behavior of the Choice Policy, we visualize the selection of action proposals during a full rollout of the dishwasher loading task (Figure~\ref{fig:choice_vis}). Each row represents one of the $K=5$ action proposals (Choice IDs), while the columns represent the progression of task phases.Interestingly, the visualization reveals that the model does not select a single head for the entire task. Instead, different proposal heads specialize in specific sub-skills. For instance, Choice ID 2 is consistently selected during the reaching and handover phases, while Choice ID 0 dominates the grasping stage. This suggests that the network naturally decomposes the long-horizon task into distinct modes, where each head learns to represent a specific portion of the behavior.

\begin{table}[!t]
\centering
\captionof{table}{Ablation study on action selection for the dishwasher task. We compare different strategies for utilizing the $K$ action proposals at inference time. \textbf{Random Choice} achieves reliable pickup but drops to 6/10 for handover and 3/10 for insertion. Both \textbf{Mean Choice} and \textbf{Single Choice} perform significantly worse, with insertion almost always failing. Our scoring-based method achieves the best results across all stages, with 10/10 for pickup, 9/10 for handover, and 7/10 for insertion. These results highlight the importance of the scoring mechanism for exploiting multi-modality.}
\setlength{\tabcolsep}{6pt}
\renewcommand{\arraystretch}{1.15}
\resizebox{\linewidth}{!}{%
\begin{tabular}{rrrr}
\toprule
Method & Pickup & Handover & Insertion \\
\cmidrule(r){1-1}
\cmidrule(r){2-4}
Random Choice & 10 / 10 & 6 / 10 & 3 / 10\\
Mean Choice & 9 / 10 & 4 / 10 & 0 / 10 \\
Single Choice (Best) & 10 / 10 & 5 / 10 & 0 / 10 \\
Single Choice (Worst) & 4 / 10 & 2 / 10 & 1 / 10 \\
Ours & 10 / 10 & 9 / 10 & 7 / 10 \\
\bottomrule
\end{tabular}
} %
\label{tab:dishwasher_ablation}
\vspace{-1em}
\end{table}

\subsection{Whole-Body Manipulation for Whiteboard Wiping}

\paragraphc{Task Setup.} In this task, a whiteboard eraser is placed at the robot’s side, outside its initial field of view. The robot must first move its head to locate the eraser and then grasp it. After pickup, the robot walks to the left toward the marked region of the whiteboard and wipes the target area clean.

This task is particularly challenging for several reasons. First, the robot’s initial pose is randomized at the start of each episode, which requires robust perception and planning. Second, locomotion introduces uncertainty because the final position after walking often deviates from the commanded target. Finally, once the robot reaches its approximate destination, it must accurately adjust its arm and body to perform effective wiping. These sources of uncertainty make the task a rigorous test of whole-body coordination. We collect 50 demonstrations for this task.

\paragraphc{Inputs.} The policy receives RGB and depth images from the head camera. Depth input is critical; we find that using RGB alone is insufficient for accurately estimating the distance to the whiteboard. Additionally, the policy receives upper-body proprioception and hand joint positions. The action space consists of target joint states for the upper body and hands, along with a low-dimensional command interface for the lower body that specifies turning, forward or backward motion, and lateral movement.

\begin{table}[!t]
\centering
\captionof{table}{Results on the loco-manipulation whiteboard wiping task. The task is divided into four stages: moving the head to locate the eraser, picking it up, walking to the board, and performing a correct wiping motion. Diffusion Policy could not be deployed because of slow inference and training instability. Both Behavior Cloning (BC) and Choice Policy reliably complete the head-movement stage (5/5). However, BC achieves higher success on pickup (2/5) and wiping (2/5), whereas Choice Policy shows limited success in these later stages. Overall success rates remain low, which highlights the difficulty of integrating perception, grasping, locomotion, and whole-body adjustment in this challenging task. Results are reported as successes out of 5 trials.}
\setlength{\tabcolsep}{4pt}
\renewcommand{\arraystretch}{1.1}
\resizebox{\linewidth}{!}{%
\begin{tabular}{rrrrr}
\toprule
Method & Move Head & Pickup & Walk & Wiping \\
\cmidrule(r){1-1}
\cmidrule(r){2-5}
Behavior Cloning & 5 / 5 & 1 / 5 & 1 / 5 & 0 / 5 \\
Choice Policy & 5 / 5 & 2 / 5 & 2 / 5 & 2 / 5 \\
\bottomrule
\end{tabular}
} %
\label{tab:wiping}
\vspace{-1.5em}
\end{table}

\paragraphc{Main Results.} The results are summarized in Table~\ref{tab:wiping}. We were unable to deploy Diffusion Policy because its slow inference and training instability. Both Behavior Cloning (BC) and Choice Policy successfully completed the initial head-movement stage. Choice Policy achieved higher success in grasping the eraser and demonstrated more reliable locomotion. In contrast, while behavior cloning occasionally reached the board, it often failed to perform stable wiping because of inaccurate final positioning.

Although the overall success rates are modest, this task is extremely challenging. It requires integrating perception, grasping, locomotion, and whole-body adjustment in a real-world environment. The results suggest that while current policies can handle individual stages reliably, achieving robust end-to-end success in long-horizon humanoid manipulation remains an open challenge.

\section{Conclusion}
\label{sec:conclusion}

In this paper, we presented a method for coordinated humanoid whole-body control for complex, long-horizon manipulation tasks. We introduced a modular teleoperation system that decomposes control into different modules. Furthermore, we proposed an efficient policy learning framework for learning from demonstrations that is capable of modeling multi-modality with a single forward pass. By utilizing the data collected through our teleoperation system and this policy learning framework, we demonstrated that humanoid robots can successfully perform several challenging manipulation tasks.

\paragraphc{Limitations and Future Work.} Although our approach shows promising results, several limitations remain. First, the visual perception component has limited generalization to substantially different scenes and objects. Incorporating more diverse training data or pre-training on large-scale datasets could improve robustness. Second, our current hand-eye coordination relies on a heuristic algorithm to direct the head toward the active hand. Developing a more adaptive or learned mechanism could further enhance performance. Addressing these limitations is a critical direction for enabling more reliable and versatile humanoid manipulation.

\section*{Acknowledgment}
This work is supported in part by the program ``Design of Robustly Implementable Autonomous and Intelligent Machines (TIAMAT)", Defense Advanced Research Projects Agency award number HR00112490425.

\bibliographystyle{IEEEtran}
\bibliography{references}

@inproceedings{zhang2025hub,
  title={HuB: Learning Extreme Humanoid Balance},
  author={Zhang, Tong and Zheng, Boyuan and Nai, Ruiqian and Hu, Yingdong and Wang, Yen-Jen and Chen, Geng and Lin, Fanqi and Li, Jiongye and Hong, Chuye and Sreenath, Koushil and Gao, Yang},
  booktitle={CoRL},
  year={2025}
}

@article{li2025reinforcement,
  title={Reinforcement learning for versatile, dynamic, and robust bipedal locomotion control},
  author={Li, Zhongyu and Peng, Xue Bin and Abbeel, Pieter and Levine, Sergey and Berseth, Glen and Sreenath, Koushil},
  journal={IJRR},
  year={2025},
}

@inproceedings{chen2024learning,
  title={Learning smooth humanoid locomotion through lipschitz-constrained policies},
  author={Chen, Zixuan and He, Xialin and Wang, Yen-Jen and Liao, Qiayuan and Ze, Yanjie and Li, Zhongyu and Sastry, S Shankar and Wu, Jiajun and Sreenath, Koushil and Gupta, Saurabh and Peng, Xue Bin},
  booktitle={IROS},
  year={2025}
}

@article{torabi2018behavioral,
  title={Behavioral cloning from observation},
  author={Torabi, Faraz and Warnell, Garrett and Stone, Peter},
  journal={arXiv preprint arXiv:1805.01954},
  year={2018}
}

@inproceedings{gu2024advancing,
  title={Advancing humanoid locomotion: Mastering challenging terrains with denoising world model learning},
  author={Gu, Xinyang and Wang, Yen-Jen and Zhu, Xiang and Shi, Chengming and Guo, Yanjiang and Liu, Yichen and Chen, Jianyu},
  booktitle={RSS},
  year={2024}
}

@article{gu2024humanoid,
  title={Humanoid-Gym: Reinforcement Learning for Humanoid Robot with Zero-Shot Sim2Real Transfer},
  author={Gu, Xinyang and Wang, Yen-Jen and Chen, Jianyu},
  journal={arXiv preprint arXiv:2404.05695},
  year={2024}
}

@inproceedings{sentis2006whole,
  title={A whole-body control framework for humanoids operating in human environments},
  author={Sentis, Luis and Khatib, Oussama},
  booktitle={ICRA.},
  year={2006},
}

@inproceedings{cheng2024open,
  title={Open-television: Teleoperation with immersive active visual feedback},
  author={Cheng, Xuxin and Li, Jialong and Yang, Shiqi and Yang, Ge and Wang, Xiaolong},
  booktitle={CoRL},
  year={2024}
}

@article{chi2023diffusion,
  title={Diffusion policy: Visuomotor policy learning via action diffusion},
  author={Chi, Cheng and Xu, Zhenjia and Feng, Siyuan and Cousineau, Eric and Du, Yilun and Burchfiel, Benjamin and Tedrake, Russ and Song, Shuran},
  journal={IJRR},
  year={2024},
}

@inproceedings{zhao2023learning,
  title={Learning fine-grained bimanual manipulation with low-cost hardware},
  author={Zhao, Tony Z. and Kumar, Vikash and Levine, Sergey and Finn, Chelsea},
  booktitle={RSS},
  year={2023}
}

@inproceedings{cheng2024expressive,
  title={Expressive whole-body control for humanoid robots},
  author={Cheng, Xuxin and Ji, Yandong and Chen, Junming and Yang, Ruihan and Yang, Ge and Wang, Xiaolong},
  booktitle={RSS},
  year={2024}
}

@inproceedings{lin2024learning,
  title={Learning Visuotactile Skills with Two Multifingered Hands},
  author={Lin, Toru and Zhang, Yu and Li, Qiyang and Qi, Haozhi and Yi, Brent and Levine, Sergey and Malik, Jitendra},
  booktitle={ICRA},
  year={2025}
}

@inproceedings{kirillov2023segment,
  title={Segment Anything},
  author={Kirillov, Alexander and Mintun, Eric and Ravi, Nikhila and Mao, Hanzi and Rolland, Chloe and Gustafson, Laura and Xiao, Tete and Whitehead, Spencer and Berg, Alexander C and Lo, Wan-Yen and Dollar, Piotr and Girshick, Ross},
  booktitle={CVPR},
  year={2023}
}

@article{simeoni2025dinov3,
  title={Dinov3},
  author={Sim{\'e}oni, Oriane and Vo, Huy V and Seitzer, Maximilian and Baldassarre, Federico and Oquab, Maxime and Jose, Cijo and Khalidov, Vasil and Szafraniec, Marc and Yi, Seungeun and Ramamonjisoa, Micha{\"e}l and Massa, Francisco and Haziza, Daniel and Wehrstedt, Luca and Wang, Jianyuan and Darcet, Timothée and Moutakanni, Théo and Sentana, Leonel and Roberts, Claire and Vedaldi, Andrea and Tolan, Jamie and Brandt, John and Couprie, Camille and Mairal, Julien and Jégou, Hervé and Labatut, Patrick and Bojanowski, Piotr},
  journal={arXiv:2508.10104},
  year={2025}
}

@inproceedings{huang2024diffuseloco,
  title={Diffuseloco: Real-time legged locomotion control with diffusion from offline datasets},
  author={Huang, Xiaoyu and Chi, Yufeng and Wang, Ruofeng and Li, Zhongyu and Peng, Xue Bin and Shao, Sophia and Nikolic, Borivoje and Sreenath, Koushil},
  booktitle={CoRL},
  year={2024}
}

@article{truong2025beyondmimic,
  title={BeyondMimic: From Motion Tracking to Versatile Humanoid Control via Guided Diffusion},
  author={Truong, Takara E and Liao, Qiayuan and Huang, Xiaoyu and Tevet, Guy and Liu, C Karen and Sreenath, Koushil},
  journal={arXiv:2508.08241},
  year={2025}
}

@article{black2024pi_0,
  title={$\pi_{0}$: A Vision-Language-Action Flow Model for General Robot Control},
  author={Black, Kevin and Brown, Noah and Driess, Danny and Esmail, Adnan and Equi, Michael and Finn, Chelsea and Fusai, Niccolo and Groom, Lachy and Hausman, Karol and Ichter, Brian and Jakubczak, Szymon and Jones, Tim and Ke, Liyiming and Levine, Sergey and Li-Bell, Adrian and Mothukuri, Mohith and Nair, Suraj and Pertsch, Karl and Shi, Lucy Xiaoyang and Tanner, James and Vuong, Quan and Walling, Anna and Wang, Haohuan and Zhilinsky, Ury},
  journal={arXiv:2410.24164},
  year={2024}
}

@article{feix2015grasp,
  title={The grasp taxonomy of human grasp types},
  author={Feix, Thomas and Romero, Javier and Schmiedmayer, Heinz-Bodo and Dollar, Aaron M and Kragic, Danica},
  journal={Transactions on Human-Machine Systems},
  year={2015},
}

@article{hsieh2025learning,
  title={Learning Dexterous Manipulation Skills from Imperfect Simulations},
  author={Hsieh, Elvis and Hsieh, Wen-Han and Wang, Yen-Jen and Lin, Toru and Malik, Jitendra and Sreenath, Koushil and Qi, Haozhi},
  journal={arXiv:2512.02011},
  year={2025}
}

@inproceedings{zhang2025flowpolicy,
  title={Flowpolicy: Enabling fast and robust 3d flow-based policy via consistency flow matching for robot manipulation},
  author={Zhang, Qinglun and Liu, Zhen and Fan, Haoqiang and Liu, Guanghui and Zeng, Bing and Liu, Shuaicheng},
  booktitle={AAAI},
  year={2025}
}

@article{guzman2012multiple,
  title={Multiple choice learning: Learning to produce multiple structured outputs},
  author={Guzman-Rivera, Abner and Batra, Dhruv and Kohli, Pushmeet},
  journal={NeurIPS},
  year={2012}
}

@inproceedings{guzman2014efficiently,
  title={Efficiently enforcing diversity in multi-output structured prediction},
  author={Guzman-Rivera, Abner and Kohli, Pushmeet and Batra, Dhruv and Rutenbar, Rob},
  booktitle={AISTATS},
  year={2014},
}

@inproceedings{lu2025mobile,
  title={Mobile-television: Predictive motion priors for humanoid whole-body control},
  author={Lu, Chenhao and Cheng, Xuxin and Li, Jialong and Yang, Shiqi and Ji, Mazeyu and Yuan, Chengjing and Yang, Ge and Yi, Sha and Wang, Xiaolong},
  booktitle={ICRA},
  year={2025},
}

@inproceedings{li2025amo,
  title={AMO: Adaptive Motion Optimization for Hyper-Dexterous Humanoid Whole-Body Control},
  author={Li, Jialong and Cheng, Xuxin and Huang, Tianshu and Yang, Shiqi and Qiu, Ri-Zhao and Wang, Xiaolong},
  booktitle={RSS},
  year={2025}
}

@article{chen2025gmt,
  title={GMT: General Motion Tracking for Humanoid Whole-Body Control},
  author={Chen, Zixuan and Ji, Mazeyu and Cheng, Xuxin and Peng, Xuanbin and Peng, Xue Bin and Wang, Xiaolong},
  journal={arXiv:2506.14770},
  year={2025}
}

@inproceedings{he2024learning,
  title={Learning human-to-humanoid real-time whole-body teleoperation},
  author={He, Tairan and Luo, Zhengyi and Xiao, Wenli and Zhang, Chong and Kitani, Kris and Liu, Changliu and Shi, Guanya},
  booktitle={IROS},
  year={2024},
}

@inproceedings{he2024omnih2o,
  title={Omnih2o: Universal and dexterous human-to-humanoid whole-body teleoperation and learning},
  author={He, Tairan and Luo, Zhengyi and He, Xialin and Xiao, Wenli and Zhang, Chong and Zhang, Weinan and Kitani, Kris and Liu, Changliu and Shi, Guanya},
  booktitle={CoRL},
  year={2024}
}

@inproceedings{he2025asap,
  title={Asap: Aligning simulation and real-world physics for learning agile humanoid whole-body skills},
  author={He, Tairan and Gao, Jiawei and Xiao, Wenli and Zhang, Yuanhang and Wang, Zi and Wang, Jiashun and Luo, Zhengyi and He, Guanqi and Sobanbab, Nikhil and Pan, Chaoyi and Yi, Zeji and Qu, Guannan and Kitani, Kris and Hodgins, Jessica and Fan, Linxi "Jim" and Zhu, Yuke and Liu, Changliu and Shi, Guanya},
  booktitle={RSS},
  year={2025}
}

@article{yang2025omniretarget,
  title={Omniretarget: Interaction-preserving data generation for humanoid whole-body loco-manipulation and scene interaction},
  author={Yang, Lujie and Huang, Xiaoyu and Wu, Zhen and Kanazawa, Angjoo and Abbeel, Pieter and Sferrazza, Carmelo and Liu, C Karen and Duan, Rocky and Shi, Guanya},
  journal={arXiv:2509.26633},
  year={2025}
}

@article{luo2025sonic,
  title={Sonic: Supersizing motion tracking for natural humanoid whole-body control},
  author={Luo, Zhengyi and Yuan, Ye and Wang, Tingwu and Li, Chenran and Chen, Sirui and Casta{\~n}eda, Fernando and Cao, Zi-Ang and Li, Jiefeng and Minor, David and Ben, Qingwei and Da, Xingye and Ding, Runyu and Hogg, Cyrus and Song, Lina and Lim, Edy and Jeong, Eugene and He, Tairan and Xue, Haoru and Xiao, Wenli and Wang, Zi and Yuen, Simon and Kautz, Jan and Chang, Yan and Iqbal, Umar and Fan, Linxi "Jim" and Zhu, Yuke},
  journal={arXiv:2511.07820},
  year={2025}
}

@article{ji2024exbody2,
  title={Exbody2: Advanced expressive humanoid whole-body control},
  author={Ji, Mazeyu and Peng, Xuanbin and Liu, Fangchen and Li, Jialong and Yang, Ge and Cheng, Xuxin and Wang, Xiaolong},
  journal={arXiv:2412.13196},
  year={2024}
}

@inproceedings{ze2024generalizable,
  title={Generalizable humanoid manipulation with 3d diffusion policies},
  author={Ze, Yanjie and Chen, Zixuan and Wang, Wenhao and Chen, Tianyi and He, Xialin and Yuan, Ying and Peng, Xue Bin and Wu, Jiajun},
  booktitle={IROS},
  year={2025}
}

@article{ze2025twist,
  title={Twist: Teleoperated whole-body imitation system},
  author={Ze, Yanjie and Chen, Zixuan and Ara{\'u}jo, Joao Pedro and Cao, Zi-ang and Peng, Xue Bin and Wu, Jiajun and Liu, C Karen},
  journal={arXiv:2505.02833},
  year={2025}
}

@inproceedings{fu2024humanplus,
  title={Humanplus: Humanoid shadowing and imitation from humans},
  author={Fu, Zipeng and Zhao, Qingqing and Wu, Qi and Wetzstein, Gordon and Finn, Chelsea},
  booktitle={CoRL},
  year={2024}
}

@article{pomerleau1988alvinn,
  title={Alvinn: An autonomous land vehicle in a neural network},
  author={Pomerleau, Dean A.},
  journal={NeurIPS},
  year={1988}
}

@inproceedings{florence2022implicit,
  title={Implicit behavioral cloning},
  author={Florence, Pete and Lynch, Corey and Zeng, Andy and Ramirez, Oscar A and Wahid, Ayzaan and Downs, Laura and Wong, Adrian and Lee, Johnny and Mordatch, Igor and Tompson, Jonathan},
  booktitle={CoRL},
  year={2022},
}

@inproceedings{shafiullah2022behavior,
  title={Behavior transformers: Cloning $k$ modes with one stone},
  author={Shafiullah, Nur Muhammad and Cui, Zichen and Altanzaya, Ariuntuya Arty and Pinto, Lerrel},
  booktitle={NeurIPS},
  year={2022}
}

@article{gu2025humanoid,
  title={Humanoid locomotion and manipulation: Current progress and challenges in control, planning, and learning},
  author={Zhaoyuan Gu and Junheng Li and Wenlan Shen and Wenhao Yu and Zhaoming Xie and Stephen McCrory and Xianyi Cheng and Abdulaziz Shamsah and Robert Griffin and C. Karen Liu and Abderrahmane Kheddar and Xue Bin Peng and Yuke Zhu and Guanya Shi and Quan Nguyen and Gordon Cheng and Huijun Gao and Ye Zhao},
  journal={arXiv:2501.02116},
  year={2025}
}

@article{li2025clone,
  title={CLONE: Closed-Loop Whole-Body Humanoid Teleoperation for Long-Horizon Tasks},
  author={Li, Yixuan and Lin, Yutang and Cui, Jieming and Liu, Tengyu and Liang, Wei and Zhu, Yixin and Huang, Siyuan},
  journal={arXiv:2506.08931},
  year={2025}
}

@article{margolis2025softmimic,
  title={SoftMimic: Learning Compliant Whole-body Control from Examples},
  author={Margolis, Gabriel B and Wang, Michelle and Fey, Nolan and Agrawal, Pulkit},
  journal={arXiv:2510.17792},
  year={2025}
}

@article{darvish2023teleoperation,
  title={Teleoperation of humanoid robots: A survey},
  author={Darvish, Kourosh and Penco, Luigi and Ramos, Joao and Cisneros, Rafael and Pratt, Jerry and Yoshida, Eiichi and Ivaldi, Serena and Pucci, Daniele},
  journal={T-RO},
  year={2023},
}

@inproceedings{ben2025homie,
  title={Homie: Humanoid loco-manipulation with isomorphic exoskeleton cockpit},
  author={Ben, Qingwei and Jia, Feiyu and Zeng, Jia and Dong, Junting and Lin, Dahua and Pang, Jiangmiao},
  booktitle={RSS},
  year={2025}
}

@inproceedings{makoviychuk2021isaac,
  title={Isaac Gym: High Performance GPU-Based Physics Simulation For Robot Learning},
  author={Makoviychuk, Viktor and Wawrzyniak, Lukasz and Guo, Yunrong and Lu, Michelle and Storey, Kier and Macklin, Miles and Hoeller, David and Rudin, Nikita and Allshire, Arthur and Handa, Ankur and State, Gavriel},
  booktitle={NeurIPS Datasets and Benchmarks},
  year={2021},
}

@inproceedings{qi2025simple,
  title={From simple to complex skills: The case of in-hand object reorientation},
  author={Qi, Haozhi and Yi, Brent and Lambeta, Mike and Ma, Yi and Calandra, Roberto and Malik, Jitendra},
  booktitle={ICRA},
  year={2025}
}

\newpage

\section*{Appendix}

\subsection{Whole-body Control Framework} 

Our whole-body control framework consists of a 20 Hz high-level command module and a 100 Hz low-level controller. The high-level command module runs on an external workstation, while the low-level controller executes on the robot’s onboard compute. Communication between the two modules is facilitated via ROS2 over Ethernet.

The high-level commands are generated from a VR controller and include head motion, arm end-effector targets, hand states, and linear and angular velocity commands for locomotion. The low-level controller is implemented as a reinforcement-learning-based locomotion policy that is capable of maintaining balance during motion.

The low-level controller is trained in IsaacGym~\cite{makoviychuk2021isaac} using the Denoising World Model Learning (DWL) approach~\cite{gu2024advancing}. Our implementation follows the training procedure and design choices described in DWL while adapting the observation space and command interface to support whole-body humanoid locomotion, as detailed in Table~\ref{tab:obs_space} and Table~\ref{tab:obs_scaling}. To enable the policy to learn both static standing and walking behaviors, we introduce an additional command sampling strategy in which static stand commands are sampled with a probability of 10\%.

\begin{table}[h]
\centering
\captionof{table}{Deployment-time observation space used by the low-level RL locomotion controller. The single-frame observation has a dimension of 49. At deployment, the policy input is formed by concatenating the observations from a rolling history buffer of \textbf{30 consecutive frames}.}
\setlength{\tabcolsep}{4pt}
\renewcommand{\arraystretch}{1.1}
\resizebox{\linewidth}{!}{%
\begin{tabular}{lcl}
\toprule
\textbf{Observation} & \textbf{Dim} & \textbf{Definition} \\
\midrule
Gait phase (sin) & 1 & $\sin(\phi)$ \\
Gait phase (cos) & 1 & $\cos(\phi)$ \\
Commanded forward velocity & 1 & $\mathrm{move}\cdot v_x$ \\
Commanded lateral velocity & 1 & $-\mathrm{move}\cdot v_y$ \\
Commanded yaw rate & 1 & $-\mathrm{move}\cdot \dot{\psi}$ \\
\midrule
Lower-body joint position & 12 & $q_\mathrm{leg}-q^{0}_\mathrm{leg}$ \\
Lower-body joint velocity & 12 & $\dot{q}_\mathrm{leg}$ \\
Previous policy action (lower limbs) & 12 & $a_{t-1}$ \\
\midrule
Base angular velocity (IMU) & 3 & $\boldsymbol{\omega}=[\omega_x,\omega_y,\omega_z]$ \\
Base orientation (Euler angles) & 3 & $\boldsymbol{\theta}=[\mathrm{roll},\mathrm{pitch},\mathrm{yaw}]$ \\
Constant bias term & 1 & $0.8$ \\
Stand flag & 1 & $1-\mathrm{move}$ \\
\bottomrule
\end{tabular}
}%
\vspace{-1em}
\label{tab:obs_space}
\end{table}

\begin{table}[h]
\centering
\captionof{table}{Observation scaling applied at deployment. All remaining observation components are used without scaling (scale = 1.0). All observation elements are clipped element-wise to $\pm 18$ before being passed to the policy.}
\setlength{\tabcolsep}{4pt}
\renewcommand{\arraystretch}{1.1}
\begin{tabular}{ll}
\toprule
\textbf{Observation} & \textbf{Scaling} \\
\midrule
Commanded forward velocity &
$\times
\begin{cases}
1.2, & v_x \ge 0 \\
0.6, & v_x < 0
\end{cases}$ \\
Commanded lateral velocity & $\times 0.3$ \\
Commanded yaw rate & $\times 0.3$ \\
Lower-body joint velocity & $\times 0.1$ \\
\bottomrule
\end{tabular}
\vspace{-3em}
\label{tab:obs_scaling}
\end{table}

\label{supp:loco}.

\end{document}